\documentclass[letterpaper, 10 pt, conference]{ieeeconf}  

\IEEEoverridecommandlockouts                              

\usepackage{multicol}
\usepackage[bookmarks=true]{hyperref}
\usepackage{caption}
\usepackage{subcaption}
\usepackage{amsfonts}
\usepackage{multirow}
\usepackage{algorithm}
\usepackage{algorithmic}
\usepackage{amsmath}
\usepackage{color}
\usepackage[normalem]{ulem}
\usepackage{graphicx}

\title{\LARGE \bf
Semantic-Based Explainable AI: Leveraging Semantic Scene Graphs and Pairwise Ranking to Explain Robot Failures}

\author{Devleena Das$^{1}$ and Sonia Chernova$^{1}$

\thanks{$^{1}$Georgia Institute of Technology, Atlanta, GA, USA
        {\tt\small \{ddas41, chernova\}@gatech.edu}}
}

\begin{document}

\maketitle
\thispagestyle{empty}
\pagestyle{empty}

\begin{abstract}
When interacting in unstructured human environments, occasional robot failures are inevitable. When such failures occur, everyday people, rather than trained technicians, will be the first to respond. Existing natural language explanations  hand-annotate contextual information from an environment to help everyday people understand robot failures. However, this methodology lacks generalizability and scalability. In our work, we introduce a more generalizable semantic explanation framework. Our framework autonomously captures the semantic information in a scene to produce semantically descriptive explanations for everyday users. To generate failure-focused explanations that are semantically grounded, we leverages both semantic scene graphs to extract spatial relations and object attributes from an environment, as well as pairwise ranking. Our results show that these semantically descriptive explanations significantly improve everyday users' ability to both identify failures and provide assistance for recovery than the existing state-of-the-art context-based explanations.

\end{abstract}

\section{Introduction}
Increasingly, robots are becoming deployed in everyday environments -- homes, hospitals, and offices -- in which the robot's primary users are everyday people rather than trained technicians \cite{zachiotis2018survey}.  Occasional robot failures are inevitable when operating in unstructured human environments, as the robot may be unable to find an object it requires, be unable to reach an object, encounter a planning error, etc.  When an error occurs, everyday people in the robot's environment are typically the first to respond, but to effectively assist in failure recovery users must have an understanding of the robot's behavior, decision making, and what went wrong \cite{das2021explainable}.


Research on Explainable AI (XAI) focuses on the development of techniques that increase the transparency and interpretability of complex, black box systems \cite{adadi2018peeking}.  The vast majority of XAI techniques developed to date have been designed for experts and system developers \cite{adadi2018peeking, ribeiro2016should, samek2019explainable, gade2019explainable}, however XAI systems also have the potential to explain the cause of a system error to everyday users. In particular, recent work has shown that natural language explanations are effective in improving user confidence in an AI system \cite{ehsan2019automated}, and in improving user assistance in fault recovery \cite{das2021explainable}.  In both of the above works, a contributing factor to the effectiveness of their explanations is the ability to incorporate \textit{situational}, or \textit{environmental} context from the agent’s environment.  However, these early works lack generalizability and scalability as both techniques require that all domain-specific contexts to be hand-annotated a priori, preventing generalization to new scenarios.  This leads to the question: \textit{How can we autonomously extract contextual information grounded in an environment to provide meaningful explanations of system failures to everyday users?}


In this work, we introduce a generalizable framework for explaining robot pick errors to non-expert users. Specifically,  we  focus  on  explaining  pick  errors that  occur  amidst  a  robot’s  task  plan,  causing  a  halt  in  the robot’s task execution. The key innovation of our approach is the use of scene graphs to produce \textit{semantically descriptive explanations} that communicate why a failure to manipulate a given object in the scene occurred. A semantic scene graph (SSG) is a data structure that represents the entities of a scene as a graph, in which objects are nodes and edges represent relationships between objects \cite{xu2020survey}. Given an image from the robot's view point of the scene, we utilize a semantic scene graph model, in conjunction with pairwise ranking \cite{furnkranz2010preference}, to produce semantically descriptive explanations. The use of scene graphs enable our approach to autonomously extract semantic context from novel scenes, thereby providing detailed explanations even for scenes and objects not previously encountered by the robot.  Additionally, we expand the types of robot failures beyond those addressed in prior work \cite{das2021explainable}.  


Our work makes the following contributions.  First, we adapt a state-of-the art semantic scene graph, MOTIFNET \cite{zellers2018neural}, to autonomously extract both inter-object spatial relations and object attribute information as contextual reasoning for robot failures in any scene. Second, we improve the semantically descriptive explanations producible through scene graphs by utilizing pairwise ranking. We show that pairwise ranking can be utilized to autonomously place attention on parts of a scene graph output that are relevant to a given failure. As a result, our framework can produce failure-focused, semantically descriptive explanations. 

We validate our approach across 4 failure types in a user study with 90 participants.  Our results show that our semantic explanation framework can produce semantically descriptive explanations that significantly improve everyday users' robot failure understanding, as well as their ability to provide assistance in failure recovery, in comparison to the state-of-the-art context-based explanations.

\section{Related Works}
\label{sec:related-works}

The XAI community has developed methodologies that increase the interpretability and transparency of black box models \cite{adadi2018peeking}. Most of these techniques are model-agnostic and aimed at understanding classification problems. Example techniques include perturbing input data to analyze consequential prediction changes \cite{ribeiro2016should, ribeiro2018anchors}, leveraging saliency maps to visualize a model's attention during prediction \cite{samek2019explainable}, and utilizing inherently interpretable models, such as decision trees or rule lists, as approximate surrogate models \cite{gade2019explainable}. While the above techniques provide insight into the inner workings of machine learning models, they have been developed for expert understanding. In our work, we focus on developing a framework that generates explanations accessible to \textit{everyday} users who are not AI experts.

Additionally, the Explainable AI Planning (XAIP) community has developed techniques that specifically focus on sequential-decision making problems \cite{chakraborti2020emerging}. Much of the work in this area has focused on generating plan explanations that explain a reasoning for the agent's selected plan. For example, these explanations may be formed by contrastive explanations that explain ``Why plan X instead of plan Y?" \cite{krarup2019model, hoffmann2019explainable}. Other works use model reconciliation, seeking to identify divergences between the mental model of the agent and the human user, to design explanations that bring such mental models closer together \cite{vasileiou2020exploiting,chakraborti2017plan}. Furthermore, when a planning problem is unsolvable, infeasible plans can be abstracted into simpler plans as a method for explaining unmet properties \cite{sreedharan2020d3wa+, sreedharan2019can}. In our work, while we generate explanations under the context of sequential-decision making tasks, our explanations focus on explaining the causes of a failure that may occur within a plan, as opposed to explaining the chosen plan.

A growing body of work is leveraging natural language explanations to explain AI decision-making to everyday users. Specifically, Ehsan et al. utilize sequence to sequence learning to autonomously generate natural language rationales that explain an agent's decision making in the context of the game Frogger \cite{ehsan2019automated}. To train their model, the authors collect annotations in the form of behavior rationales from everyday users.  Their results demonstrate that users significantly prefer ``complete-view" rationales, which utilize the entire state space as context, as opposed to ``focused-view", which utilize only a subset of the full state space.
Most closely related to our work, Das et al. utilize sequence to sequence learning to autonomously generate natural language explanations in the context of robot failures \cite{das2021explainable}. To train their model, the authors collect expert annotations for each timestep in a robot's task plan.  Their results demonstrate that the inclusion of environmental context and history of past actions help improve user ability to correctly identify failures and their solutions. In our work, we aim to produce natural language explanations grounded in semantic context. Instead of hand-annotating contextual information, we leverage semantic scene graphs to autonomously capture the semantic information from a scene. In doing so, we are able to expand the set of explainable failure scenarios from \cite{das2021explainable}. 
 

Within the robotics community, scene graphs have been utilized for scene analysis and goal-directed manipulation. For instance, Zeng et al. use scene graphs to parse a goal scene configuration in efforts to allow robots to efficiently motion plan and transform an initial scene into such pre-defined goal \cite{zeng2018semantic}. Sui et al. leverage scene graphs for axiomatic particle filtering, which allows robots to disambiguate objects in a cluttered scene for effective manipulation \cite{sui2017goal}. Kenfack et al. intersect Visual Question Answering (VQA) and robotics and develop a robotVQA architecture. RobotVQA provides semantically-grounded answers to questions about a scene with the motivation to illicit more meaningful robot object manipulations in the future \cite{kenfack2020robotvqa}. Scene graphs have also been utilized to mitigate safety risks in human-robot-collaboration scenarios \cite{riaz2020scene,inam2018risk}. 
Most closely related to our work, scene graphs have been shown to be effective in producing explainable answers for VQA queries \cite{ghosh2019generating}. 
The authors utilize attention maps to autonomously select relevant relations from a scene graph to explain their VQA model's answers.
In our work, instead of attention maps, we utilize pairwise ranking of inter-object relations and object attributes to provide ranked, semantically descriptive explanations which include only the relevant semantic information needed to explain a robot failure.


Research on fault diagnosis has led to the development of a wide range of techniques for diagnosing, and suggesting recoveries for robot failures \cite{khalastchi2018fault}. Example techniques include execution monitoring \cite{banerjee2020taking,ingrand2017deliberation}, sensor-processing \cite{jager2014assessing}, neural networks \cite{vemuri1998neural,van2015robust}, and statistical filtering \cite{verma2004real}. However, such techniques are either aimed at autonomous recovery \cite{khalastchi2018fault,banerjee2020taking}, or at aiding an expert operator who is deeply familiar with the inner workings of the system, not everyday users.


\begin{figure*}[t]
\centering
\begin{subfigure}[b]{0.242\textwidth}
  \centering
  \includegraphics[width=\textwidth]{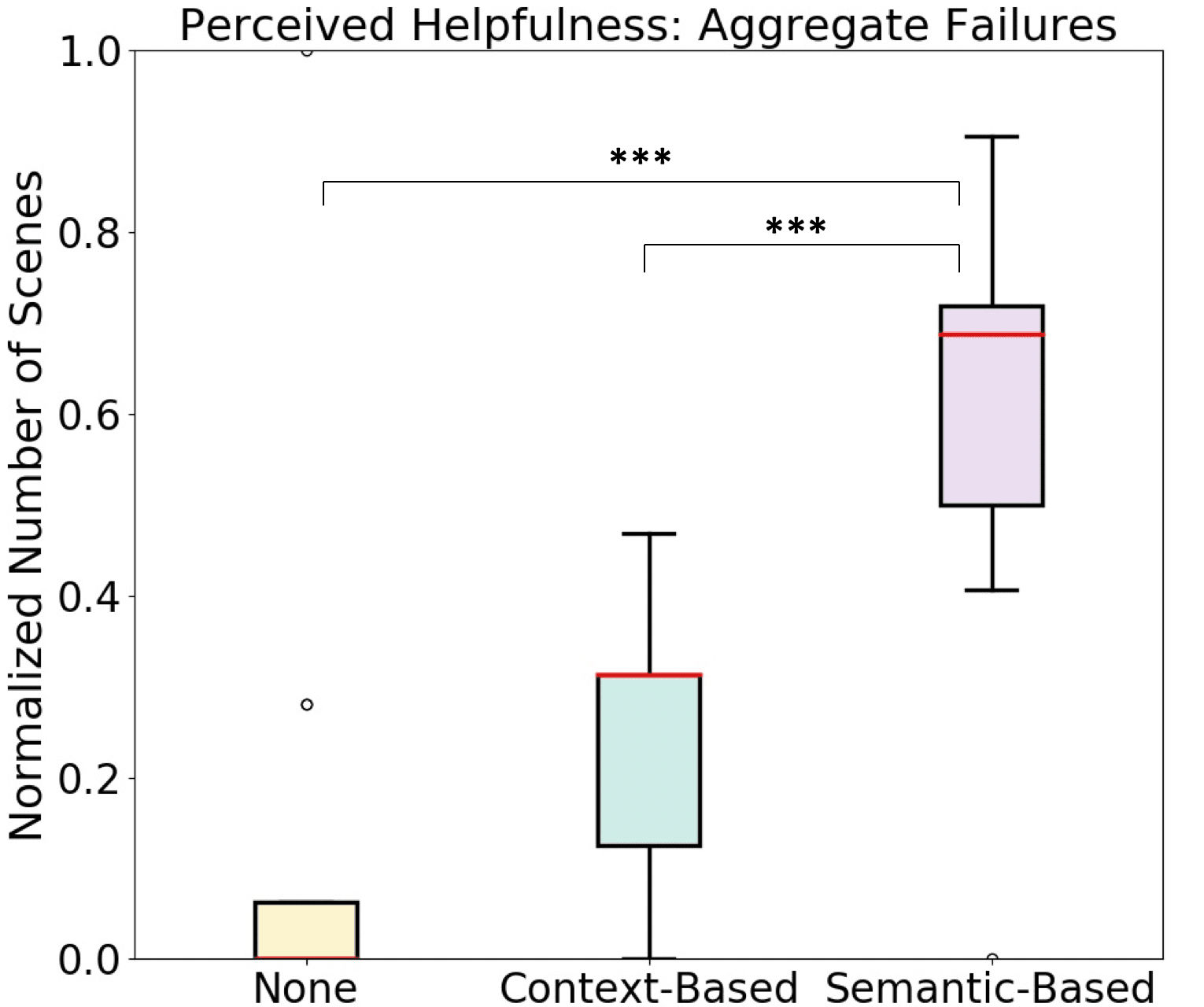}
  \caption[]%
  {{}}
\end{subfigure}\quad
\begin{subfigure}[b]{0.24\textwidth}
  \centering
  \includegraphics[width=\textwidth]{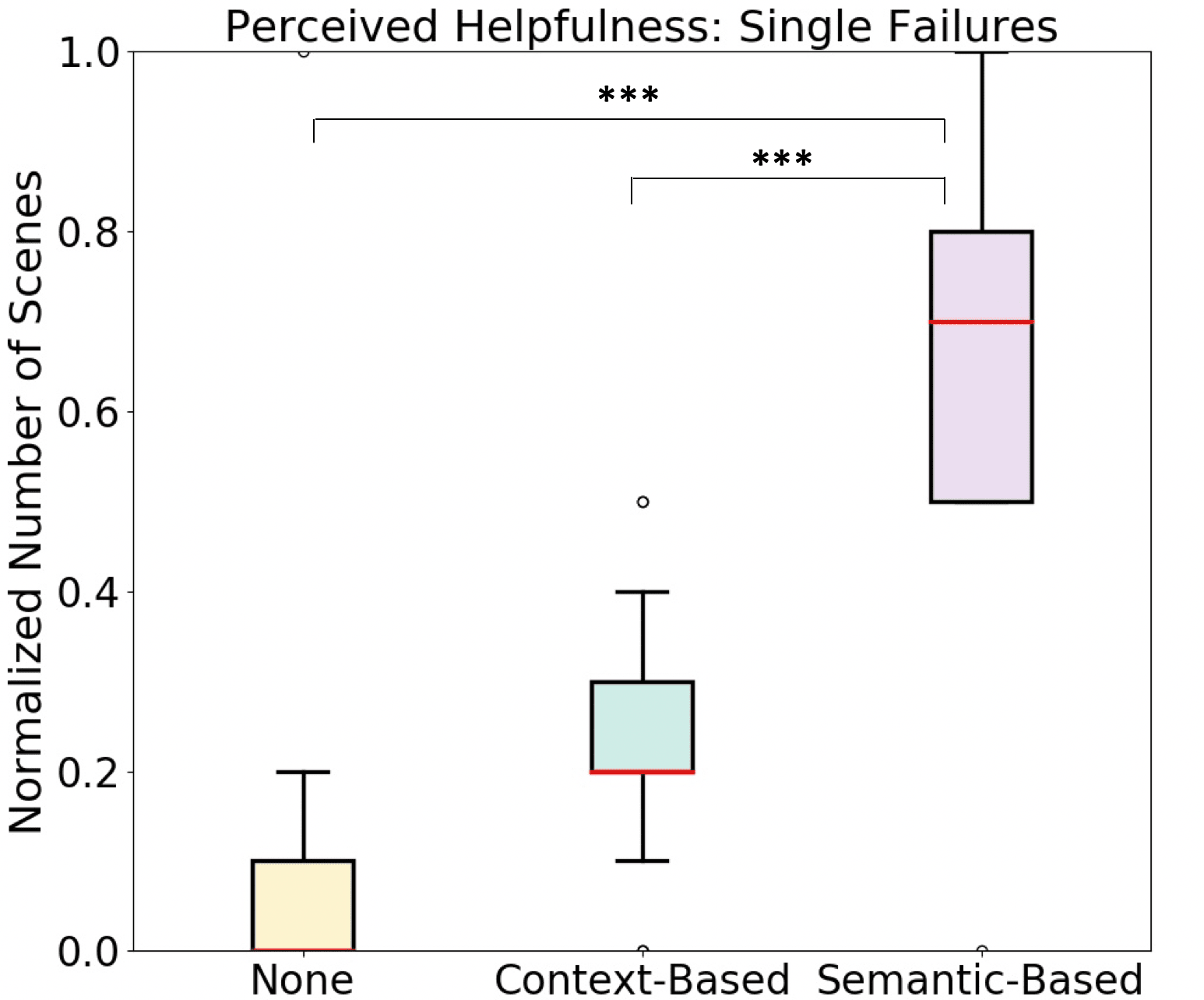}
    \caption[]%
     {{}}
     \end{subfigure}
\begin{subfigure}[b]{0.24\textwidth}
  \centering
  \includegraphics[width=\textwidth]{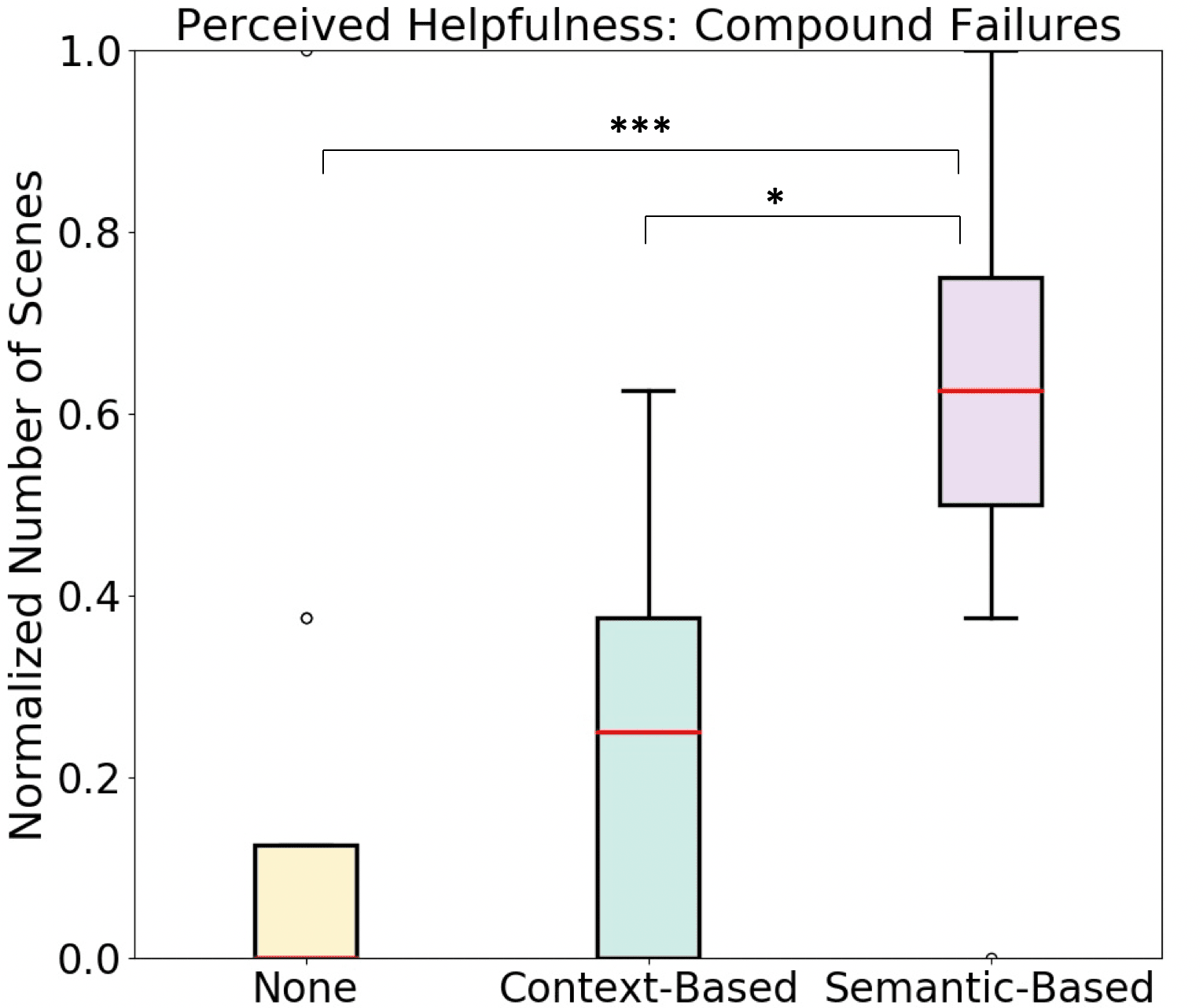}
    \caption[]%
     {{}}
     \end{subfigure}
\begin{subfigure}[b]{0.233\textwidth}
  \centering
  \includegraphics[width=\textwidth]{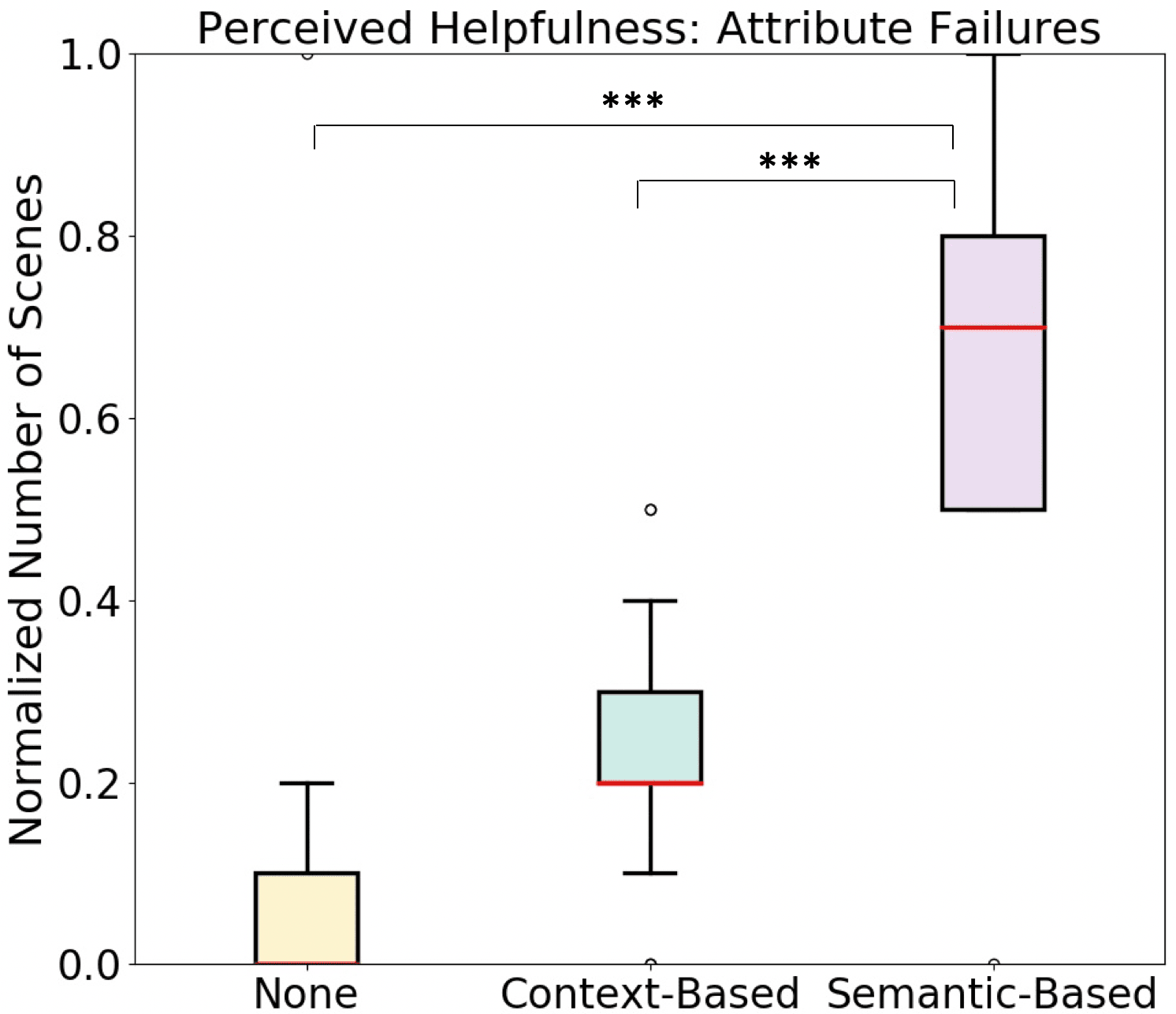}
    \caption[]%
     {{}}
     \end{subfigure}
\caption{Average selected explanation type based on users' perceived helpfulness. Statistical significance is reported as: * p $<$ 0.05, ** p $<$ 0.01, *** p$<$ 0.001.}
\label{fig:perceived-qual}
\end{figure*}

\section{Semantically Descriptive Explanations}
\label{sec:section-3}

\begin{figure}
\includegraphics[width=0.9\columnwidth]{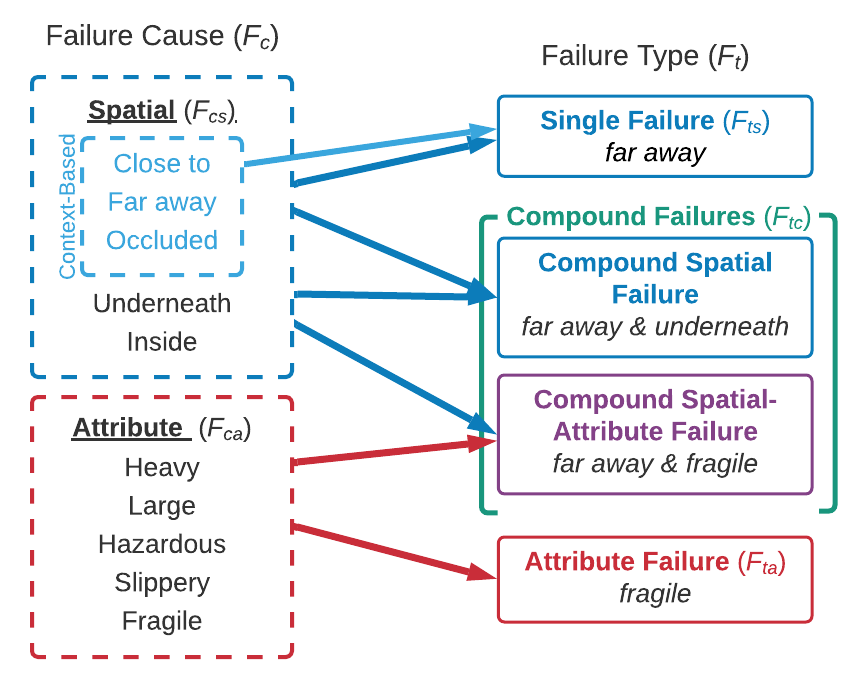}
\caption{Breakdown of failure causes, $F_{c}$ and failure types $F_{t}$ expanded in this work, as well as the specific $F_{c}$ and $F_{t}$ explained by CB explanations from prior work. }
\label{fig:failure_types}
\end{figure}

Given a set of failure types, $F_{t}$, that prevent a robot from picking up a desired object, $d_{obj}$, our goal is to produce natural language explanations that help everyday users (1) understand the cause of the robot's failure, and (2) identify the correct way to assist the robot in recovery. As seen in Figure \ref{fig:failure_types}, we characterize $F_{t}$ by $\{F_{ts}, F_{tc}, F_{ta}\}$. In this set $F_{ts}$ represent single spatial failures, $F_{tc}$ represent compound spatial failures, and $F_{ta}$ represent attribute failures. We denote that each failure type can be caused by the set $F_{c} = \{F_{cs}, F_{ca}\}$, where $F_{ca}$ define failures caused by spatial relationships and $F_{cs}$ define object attribute causes. We believe that an effective solution to our objective is to utilize the semantic information from a scene and generate semantically descriptive explanations for robot failures. We qualify a semantically descriptive explanation as one that utilizes the inter-object spatial relationships and object specific attributes from a given scene. An inter-object spatial relationship describes an object's location with respect to other objects in a scene. For instance, ``a credit card is \textit{underneath} a newspaper". An object specific attribute describes a property of the object, such as ``the vase is \textit{fragile}".


To validate the importance of semantically descriptive explanations, we developed a qualitative study in which our explanations were derived from hand-crafted semantic relationships in a scene\footnote{Participants were recruited through Amazon Mechanical Turk; They were 18 years or older (M=31.4 SD=9.5) and were compensated \$2.50 for the task.}. We compared our approach to the only previously published error explanation technique \cite{das2021explainable}.  In \cite{das2021explainable}, context-based (CB) explanations were generated for novel scenes based on similarity to previously annotated scenarios. CB explanations focus on single failure types, $F_{ts}$, and are developed without the use of semantic information; for instance, in the example where a credit card is underneath a newspaper, a CB explanation would be the ``credit card is occluded".  Although correct, this statement is more vague than the one that utilizes the semantic scene information.  

In our study, each user was presented with a scene from a household environment, as well as three descriptions for the cause of a pick error: no explanation, a CB explanation, and a semantically descriptive explanation.  Users were asked to select which explanation was most helpful in understanding the robot's cause of failure. In Figure~\ref{fig:perceived-qual}, we present participants' perceived usefulness both over aggregated failures, as well as across each failure type. We observe that for all failure types, the semantically descriptive explanations are perceived as significantly more useful than no explanations and the CB explanations.

\begin{figure*}
\centering
\includegraphics[width=15cm]{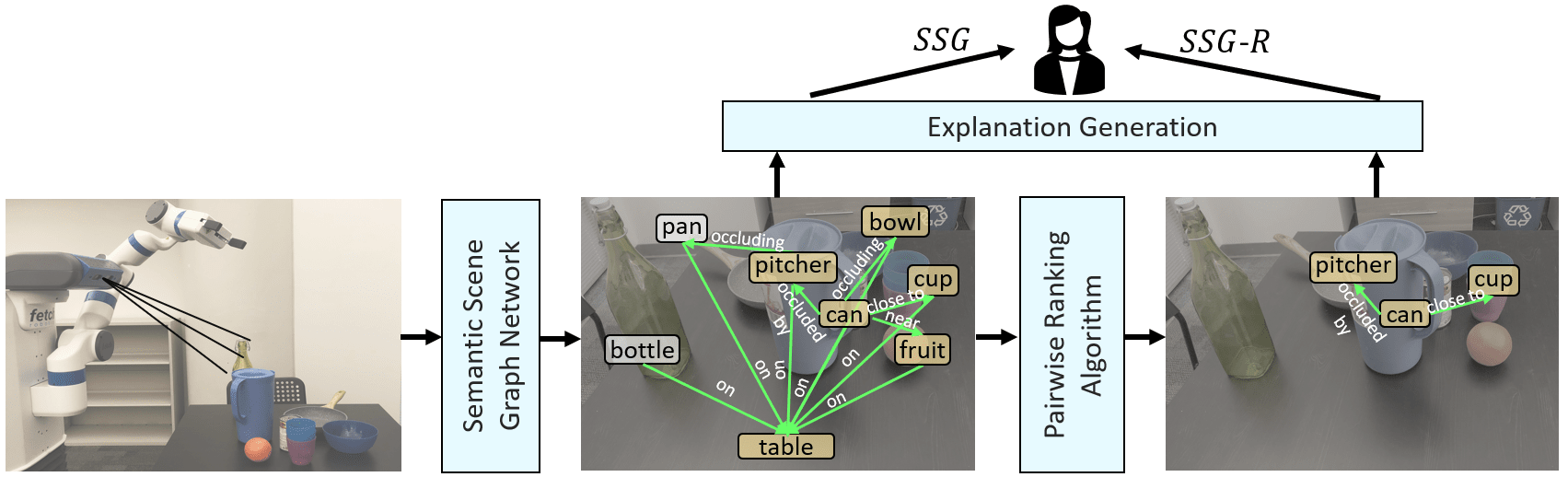}
\caption{The semantic explanation framework is used to generate unranked $SSG$ explanations and ranked $SSG$-$R$ explanations. The framework consists of three modules: (1) a scene graph network that autonomously extracts semantic information from a scene, (2) pairwise algorithm that ranks semantic information based on relevancy to a failure scenario, and (3) an explanation generation template that produces both variants of natural language explanations.}
\label{fig:sys_overview}
\end{figure*}

\begin{figure}
\centering
\includegraphics[width=0.7\columnwidth]{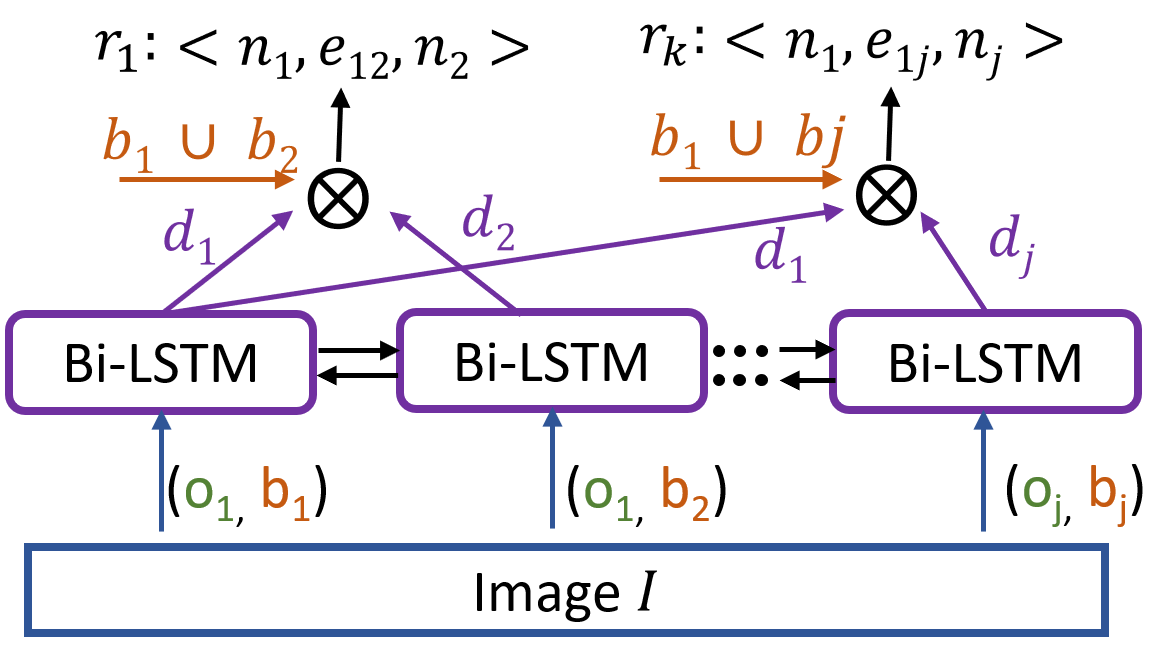}
\caption{Our adapted MOTIFNET model architecture utilized to evaluate predicate classification.}
\label{fig:model_arch}
\end{figure}

\begin{figure}[t]
\centering
\includegraphics[width=0.9\columnwidth]{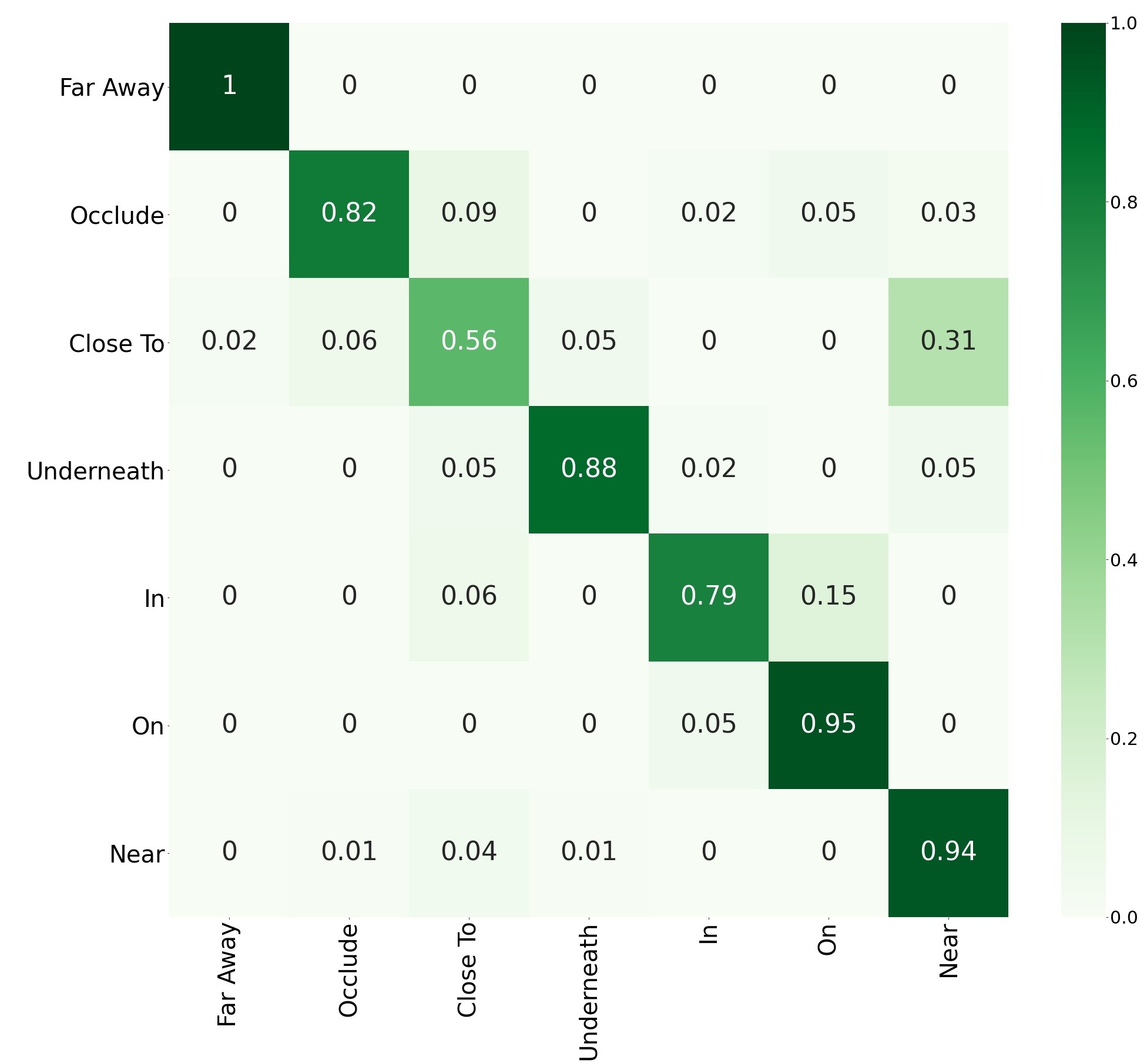}
\caption{Confusion Matrix of our SSG model's performance where the y-axis denote the ground truth predicates and the x-axis denote predicted predicates.}
\label{fig:cm}
\end{figure}

\section{Scene Graph Model}
\label{sec:ssg}

Given that explanations grounded in inter-object relationships and object attributes were perceived as significantly more useful by everyday users, we next developed a methodology to autonomously generate these semantically descriptive explanations. To do so, we introduce the semantic explanation framework shown in Figure \ref{fig:sys_overview}. Our framework leverages semantic scene graphs and pairwise ranking to deliver two variants of semantically descriptive explanations: $SSG$ and $SSG$-$R$. In Section \ref{sec:ssg_model} we discuss how we adapt the semantic scene graph architecture MOTIFNET \cite{zellers2018neural} to predict spatial relationships and object attribute from a given scene. In Section \ref{exp-ssg} we demonstrate how the semantic scene graph model outputs are utilized to template \textit{unranked} $SSG$ explanations. We also showcase the utility of pairwise ranking to develop \textit{ranked} $SSG$-$R$ explanations that only select the relevant semantic information in a scene. Finally, in Section \ref{sec:quant-eval}, we perform a quantitative analysis comparing both variants of semantically descriptive explanations with the existing baselines. We demonstrate the effectiveness of \textit{ranked} $SSG$-$R$ to everyday users in improving their understanding of robot failures and ability to accurately assist in failure recovery.

\subsection{Semantic Scene Graphs}
\label{sec:ssg_model}
A scene graph, $G$, describes the semantic information contained in a given image and is represented by a set of nodes, $N$, and edges, $E$, \cite{zellers2018neural}. Each $n_{i} \in N$ is defined by a pair $(b_{i},o_{i})$ in which $b_{i}$ represents a detected bounding box, and $o_{i}$ represents the associated object label. Similar to \cite{armeni20193d}, we also provide each $n_{i}$ with an object attribute, $a_{i} \in A$, where $A$ is the set of object attributes from Figure \ref{fig:failure_types}. Additionally, each $e_{ij} \in E$ is defined as a predicate label between $n_{i}$ and $n_{j}$. The predicate labels refer to the inter-object relations in a scene (e.g., underneath, inside, close to). Given these definitions, the output of a scene graph is defined by a set of triples $R$ = \{$r_{1}, r_{2}...r_{m}\}$ in which each $r_{k} \in R$ is defined by $<n_{i}, e_{ij}, n_{j}>$. 


We adapt the state-of-the-art scene graph model MOTIFNET \cite{zellers2018neural}\footnote{We utilize the codebase provided by \cite{tang2020sggcode} to adapt our MOTIFNET.} to predict spatial relationships and object attributes in a given scene. 
Figure~\ref{fig:model_arch} depicts our model architecture.
For the purposes of our application, we evaluate our model on predicate classification, a form of SSG evaluation that utilizes both ground truth bounding box regions and object labels to predict predicate edge labels. As shown in Figure~\ref{fig:model_arch}, ground truth bounding box regions, $\{b_{1}, b_{2}..b_{j}\}$, and object labels, $\{o_{1}, o_{2}..o_{j}\}$, are extracted from an image $I$ and passed into a bi-LSTM network structure with highway connections \cite{srivastava2015training}. To predict a triple $r_{k}$, the contextualized information for two objects, $o_{i}$ and $o_{j}$, is utilized in conjunction with the union of corresponding bounding box information, $b_{i}$ and $b_{j}$, to determine the final predicate label $e_{ij}$.


\subsection{Data Collection}
\label{sec:ssg_data_collection}
To train our adapted MOTIFNET model, we collected a dataset $D$ consisting of 188 household cluttered images from the AI2Thor simulator \cite{kolve2017ai2}. Images in $D$ were taken from the perspective of the robot, and capture the unstructured, cluttered environment of human households. The images in $D$ represent what a robot would perceive as it attempts to pick up a desired object $d_{obj}$. Therefore, our images capture  close-view scenes of major receptacles such as kitchen countertops, dining tables or desks. On average, each image in $D$ includes 13 objects. These objects include  approximately 6 object attributes and 30 inter-object spatial relations. In each image, we assume every object to have only one attribute, including ``None" which denotes when an object does not include an attribute listed in Figure~\ref{fig:failure_types}.

\subsection{Training \& Evaluation}
\label{sec:model-training}
We train our adapted MOTIFNET on ground truth bounding box regions and object labels to predict predicate and attribute labels. We utilize a 66\%-17\%-17\% split in which we use 126 scenes for training, 32 for validation and 32 for evaluation. Our model is trained with 2000 iterations and utilizes a Cross Entropy loss that is optimized using SGD with a learning rate of 0.01 and momentum of 0.9. 

The confusion matrix in Figure~\ref{fig:cm} shows the average performance of our predicate classification. 
Our adapted MOTIFNET can generalize the predicate labels with 84.9\% accuracy. While our model has low false positive labels for most relations, we see that our model has a greater challenge differentiating labels that are semantically similar. For example, ``close to" is most erroneously classified as ``near". Similarly, ``in" is most erroneously classified as ``on". These labels learn the relationship between 2D bounding boxes, with a threshold as a discriminator. Improvements on the SSG model architecture, as well as additional training data, will likely lead to improvements in the classification accuracies.  As we will show in Section \ref{sec:quant-eval}, the current level of performance is sufficient in effectively conveying error explanations to users.


\begin{figure*}[t]
\centering
\includegraphics[width=15cm]{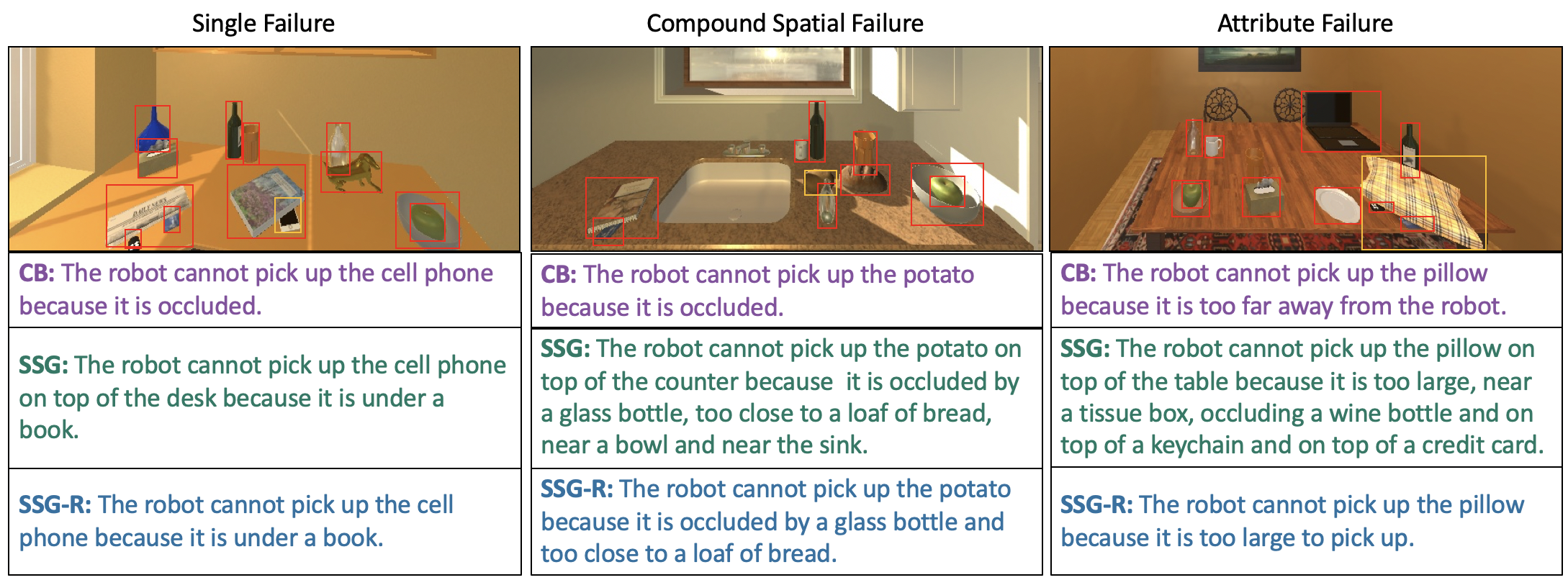}
\caption{Sample failure scenarios, where the red boxes indicate the bounding boxes of ground truth objects in the scene, and the yellow box represents $d_{obj}$. We illustrate model-generated explanations, comparing CB, $SSG$ and $SSG$-$R$ explanations.}
\label{fig:explanation_types}
\end{figure*}

\section{Generating Explanations from SSGs}
\label{exp-ssg}
Given a desired object $d_{obj}$, an image $I$ of the local environment from the robot's camera corresponding to one of the failures in Figure \ref{fig:failure_types}, and the corresponding image scene graph $G$, our goal is to produce semantically descriptive, natural language explanations that reason about why a robot cannot pick up $d_{obj}$. Below, we detail how explanation variants $SSG$ and $SSG$-$R$ are generated.

\subsection{$SSG$ Explanations}
To develop $SSG$ explanations, we follow a template-based approach that traverses a scene graph, $G$, and extracts a subgraph $g$ containing all relations $r_k \in R$ which contain $d_{obj}$ as a node in the triple. To generate an SSG explanation, we describe $g$, the elements of the scene that pertain to our object of interest.  Specifically, we format the explanation as \textit{The robot could not pick up the $<d_{obj}>$ because $<reasoning>$}, where $<reasoning>$ is a list of phrases that enumerates all of the object relations $r_k$.



In Figure \ref{fig:explanation_types} we showcase examples of $SSG$ explanations in the context of our failure types $F_{t}$. In every scene, the $SSG$ explanations include \textit{all} relationships associated with $d_{obj}$. We observe that $SSG$ explanations are more semantically richer, and detailed than their CB counterpart. However, we also observe that these explanations include extraneous, semantic information that hides the true cause of a failure. In other words, a drawback of these $SSG$ explanations is that the scene graph model has no insight into \textit{which} $r_{k} \in g$ are relevant in describing the robot's failure. 



\subsection{$SSG$-$R$ Explanations via Pairwise Ranking}
To provide only relevant relations as $reasoning$ for a failure, we develop $SSG$-$R$ explanations. In addition to extracting a subgraph $g$, we utilize pairwise ranking to autonomously determine the relevancy of each triple $r \in g$. Pairwise ranking is used to learn preferences between pairs of entities when multiple available entities exist \cite{furnkranz2010preference}. In our application, a preference denotes how accurately a relationship describes the true cause(s) of failure(s).

To formulate our pairwise ranking problem we let a pair of relationship triples, $r_{k}$ and $r_{m}$, be defined by feature vectors $f_{k} = [{e_{ij}}_{k}, {a_{i}}_{k}, {a_{j}}_{k}]$ and $f_{m} = [{e_{ij}}_{m}, {a_{i}}_{m}, {a_{j}}_{m}]$. Recall from Section \ref{sec:ssg} that $e{ij}$ represents the predicate label between two object nodes $n_{i}$ and $n_{j}$, while $a_{i}$ and $a_{j}$ represent the predicted object attributes for $n_{i}$ and $n_{j}$. 

Given our feature vectors, Algorithm 1 further details the pairwise ranking process for a subgraph $g$. We utilize $nLabels$ to determine how many binary classifiers are required to represent each unique pair of labels. For our application, we had a set of three preference labels $\{0,1,2\}$, and therefore required three binary classifiers to be instantiated, one for each label pair: $\{0,1\}$, $\{0,2\}$, and $\{1,2\}$. The list $classifierList$ contains the three instantiated classifiers (lines 1-2). We iterate through $classifierList$, and pass the feature vectors of each relationship pair, $f_{k}$ and $f_{m}$, as an input to each classifier (lines 3-4). The following logic denotes how a predicted label $y_{km}$ is determined for a given input $[f_{k}, f_{m}]$:
$$
y_{km} = \begin{cases}
0, & \text{if $r_{k} > r_{m}$} \\ 
1, & \text{if $r_{k} < r_{m}$} \\ 
2, & \text{if $r_{k} = r_{m}$}
\end{cases}
$$
A predicted label 0 represents when the relationship $r_{k}$ better describes the cause of failure than $r_{m}$. A label 1 represents when the relationship $r_{m}$ better describes the cause of failure than $r_{k}$, and a label 2 represents when both relationships equally describe the cause of failure. For our purposes, we utilize random forest classifiers, trained using cross validation via scikit-learn. Depending on the predicted label, the rank of one or both relationships is incremented (lines 5-10). The resulting list of relationships $L_{r}$, sorted by rank, is returned by the algorithm (line 14-15). Note, annotation of a training label 0, 1 or 2 is determined via domain knowledge of the failure scenario. 


To develop an $SSG$-$R$ explanation, we follow the identical template utilized for $SSG$; however, $reasoning$ is now replaced with the top ranked relationship(s) in $L_{r}$. Note, that including a tie label 2, that represents equally ranked relationships, allows our pairwise ranking to represent more than one relationship with the max rank. Figure \ref{fig:explanation_types} exemplifies how pairwise ranking can eliminate the presence of extraneous relationships when compared to $SSG$ explanations, while still being semantically richer than CB explanations.

\begin{algorithm}[h!] 
\caption{Pairwise Ranking Algorithm}  
\textbf{Input:}{~$g$ - scene subgraph} \\ 
\textbf{Output:}{~$L_{r}$ - ranked relation list } \\
\begin{algorithmic} [1]
\STATE $nLabels$ = 3
\STATE classifierList = loadClassifiers($nLabels\cdot$($nLabels$-1)/2) \\ \medskip
\FORALL{$f_{k}$, $f_{m}$ in $g ~\|~m \neq k$}
\FOR{classifier in classifierList}
\IF{classifier([$f_{k}$, $f_{m}$]) = 0}
\STATE {incrementRank($f_{k}$)}
\ELSIF{classifier([$f_{k}$, $f_{m}$]) = 1}
\STATE {incrementRank($f_{m}$)}
\ELSIF{classifier([$f_{k}$, $f_{m}$]) = 2}
\STATE {incrementRank($f_{k}$, $f_{m}$)}
\ENDIF
\ENDFOR
\ENDFOR
\STATE $L_{r}$ = sortByRank($g$)
\RETURN  $L_{r}$ 
\end{algorithmic}
\end{algorithm}

\begin{figure*}[t]
\centering
\begin{subfigure}[b]{0.24\textwidth}
  \centering
  \includegraphics[width=\textwidth]{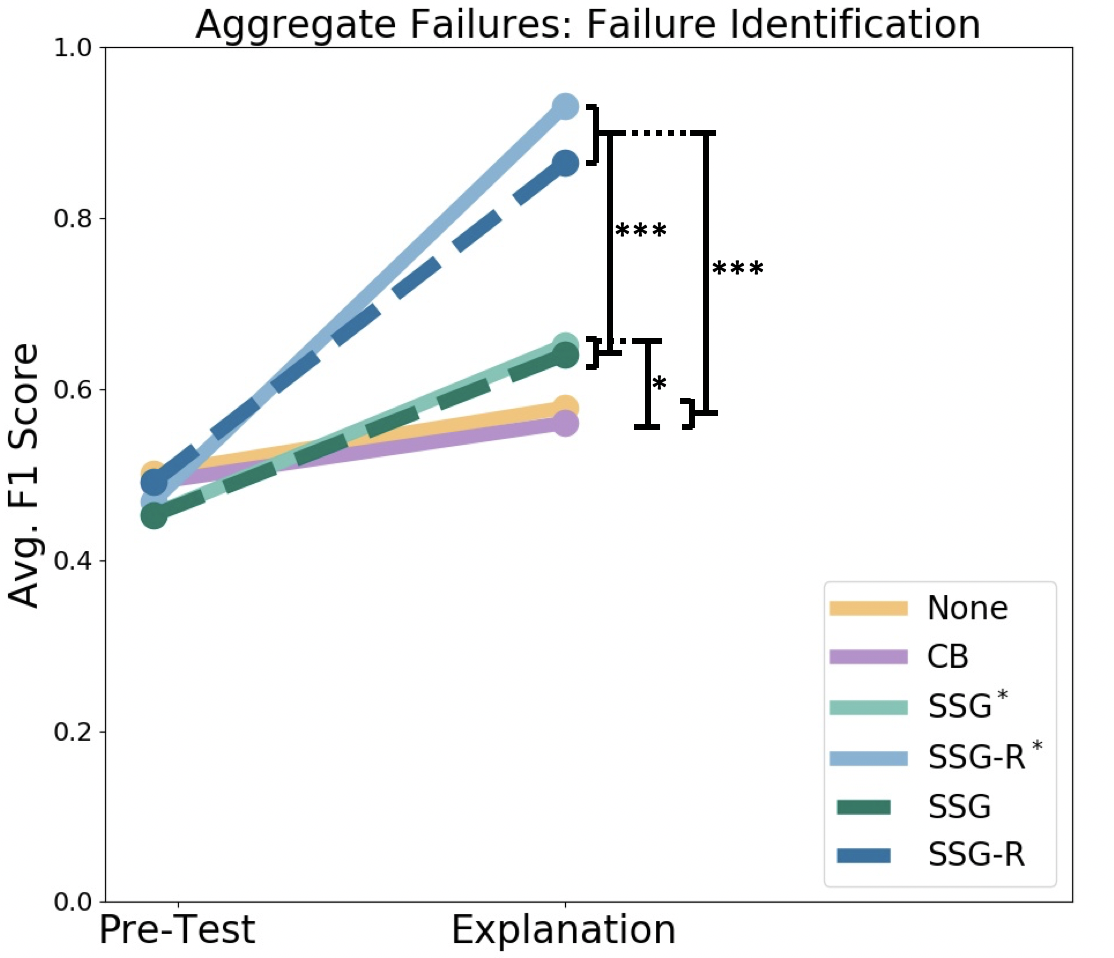}
  \caption[]%
  {{}}
\end{subfigure}\quad
\begin{subfigure}[b]{0.24\textwidth}
  \centering
  \includegraphics[width=\textwidth]{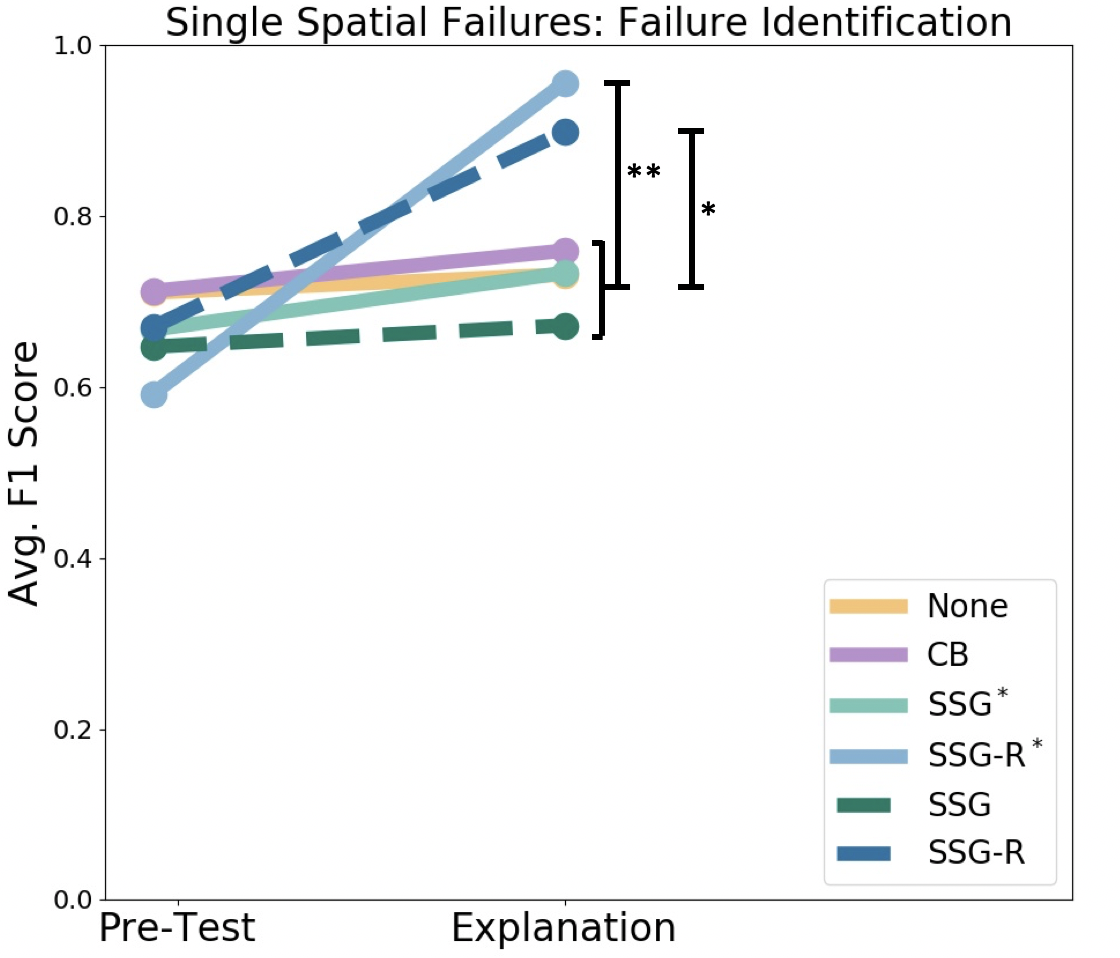}
    \caption[]%
     {{}}
     \end{subfigure}
\begin{subfigure}[b]{0.24\textwidth}
  \centering
  \includegraphics[width=\textwidth]{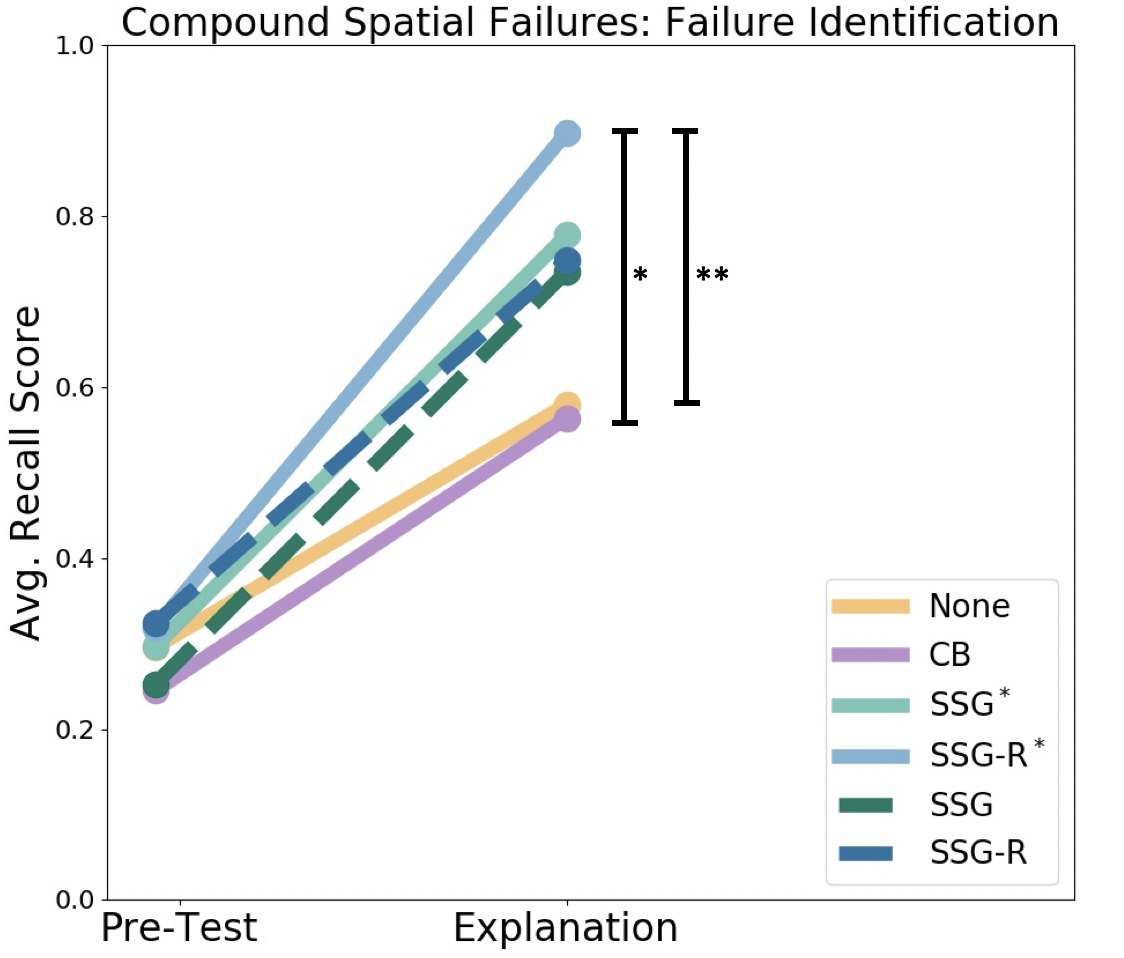}
    \caption[]%
     {{}}
     \end{subfigure}
\begin{subfigure}[b]{0.233\textwidth}
  \centering
  \includegraphics[width=\textwidth]{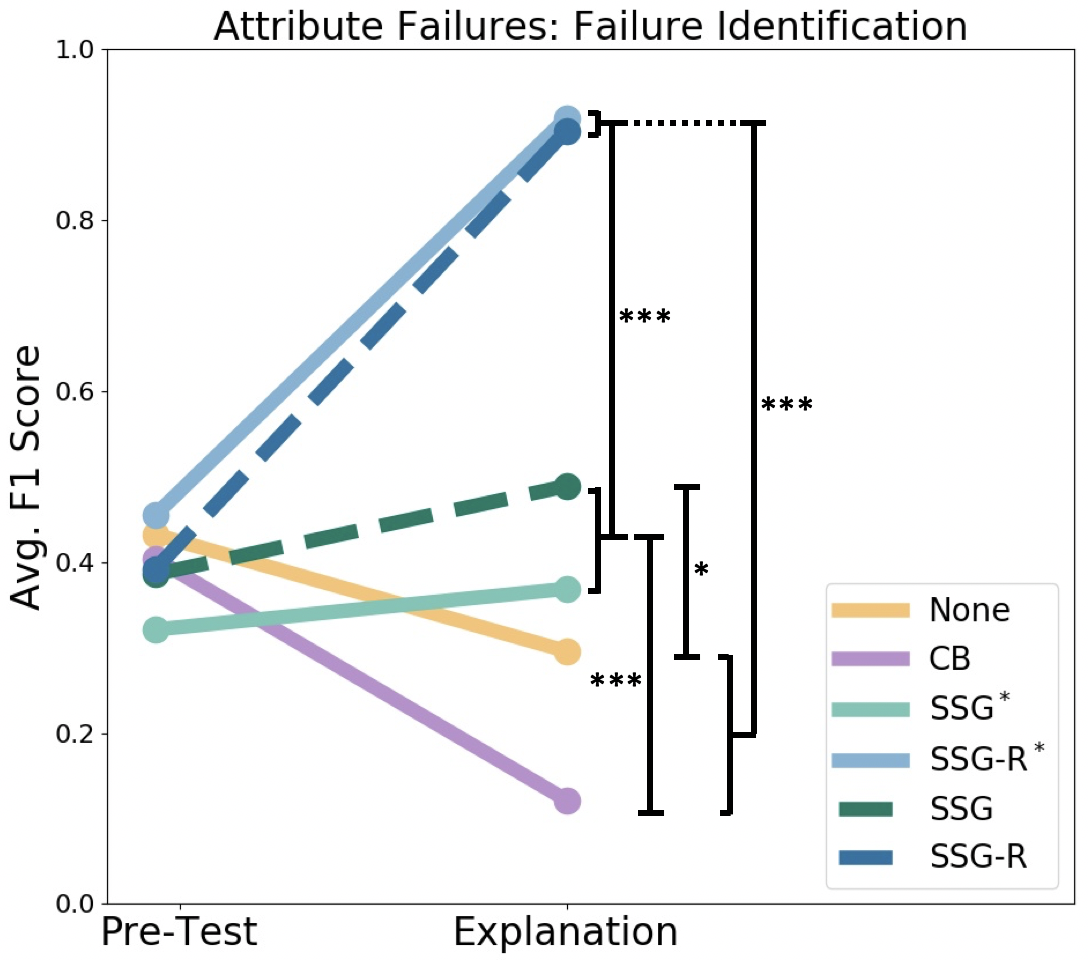}
    \caption[]%
     {{}}
     \end{subfigure}
\caption{Average F1 and Recall score for participants' failure identification across  all study conditions. Statistical significance is reported as: * p $<$ 0.05, ** p $<$ 0.01, *** p$<$ 0.001. }
\label{fig:q1-qual}
\end{figure*}

\begin{figure*}[t]
\centering
\begin{subfigure}[b]{0.24\textwidth}
  \centering
  \includegraphics[width=\textwidth]{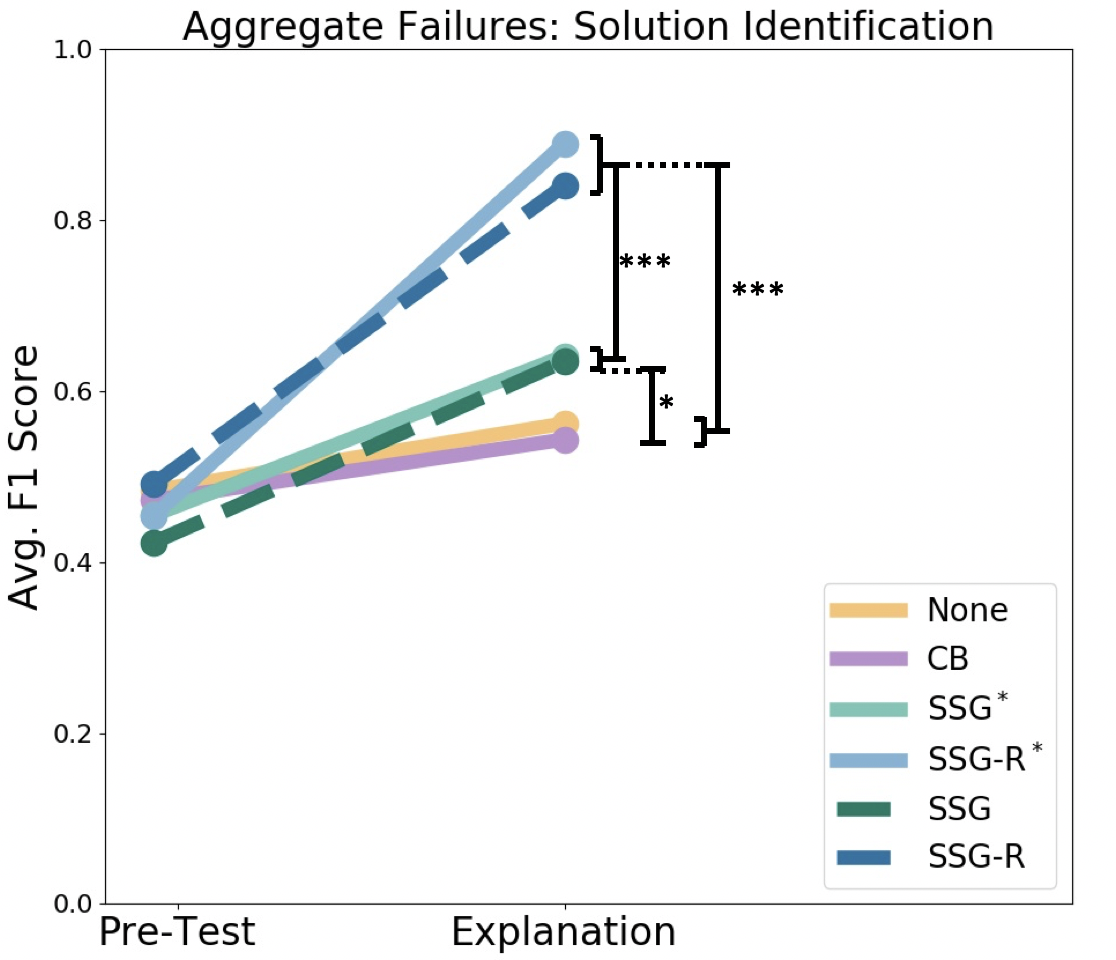}
  \caption[]%
  {{}}
\end{subfigure}\quad
\begin{subfigure}[b]{0.24\textwidth}
  \centering
  \includegraphics[width=\textwidth]{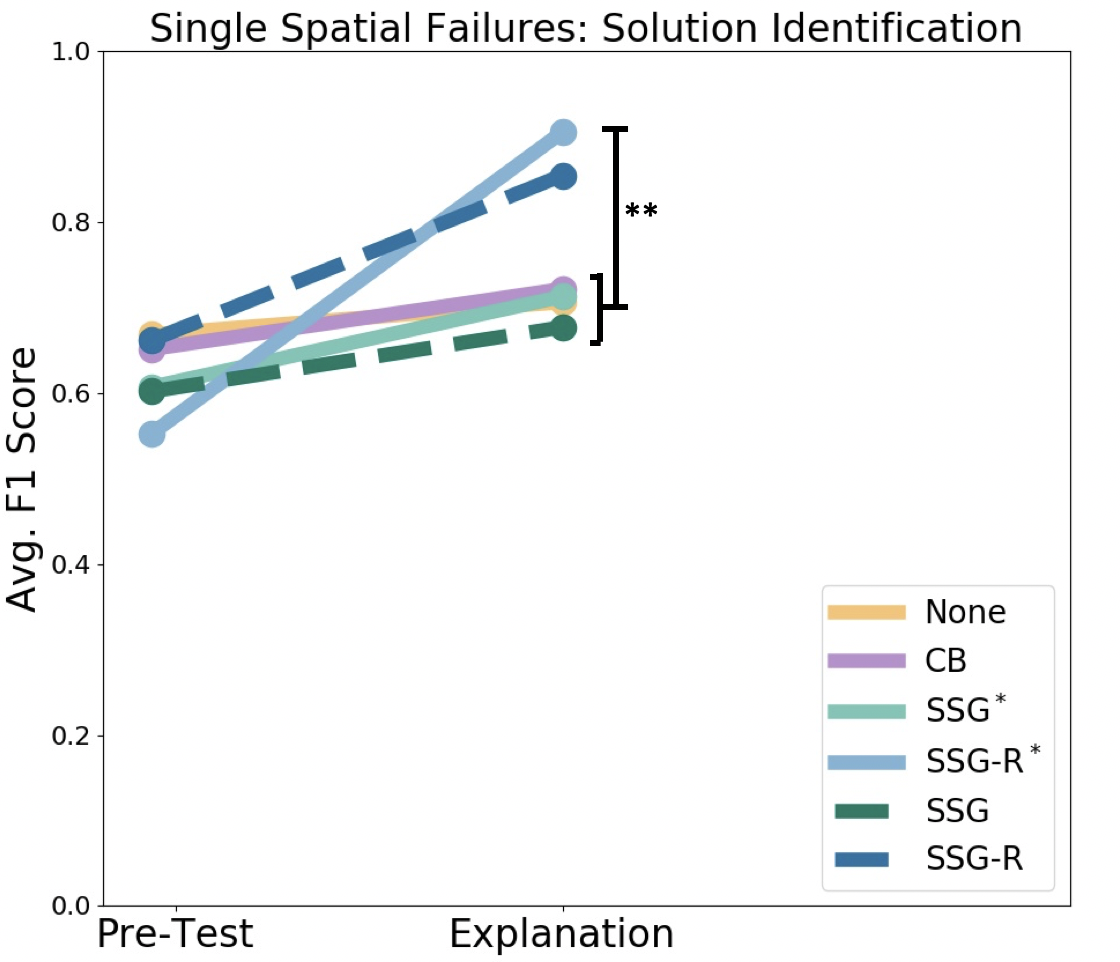}
    \caption[]%
     {{}}
     \end{subfigure}
\begin{subfigure}[b]{0.24\textwidth}
  \centering
  \includegraphics[width=\textwidth]{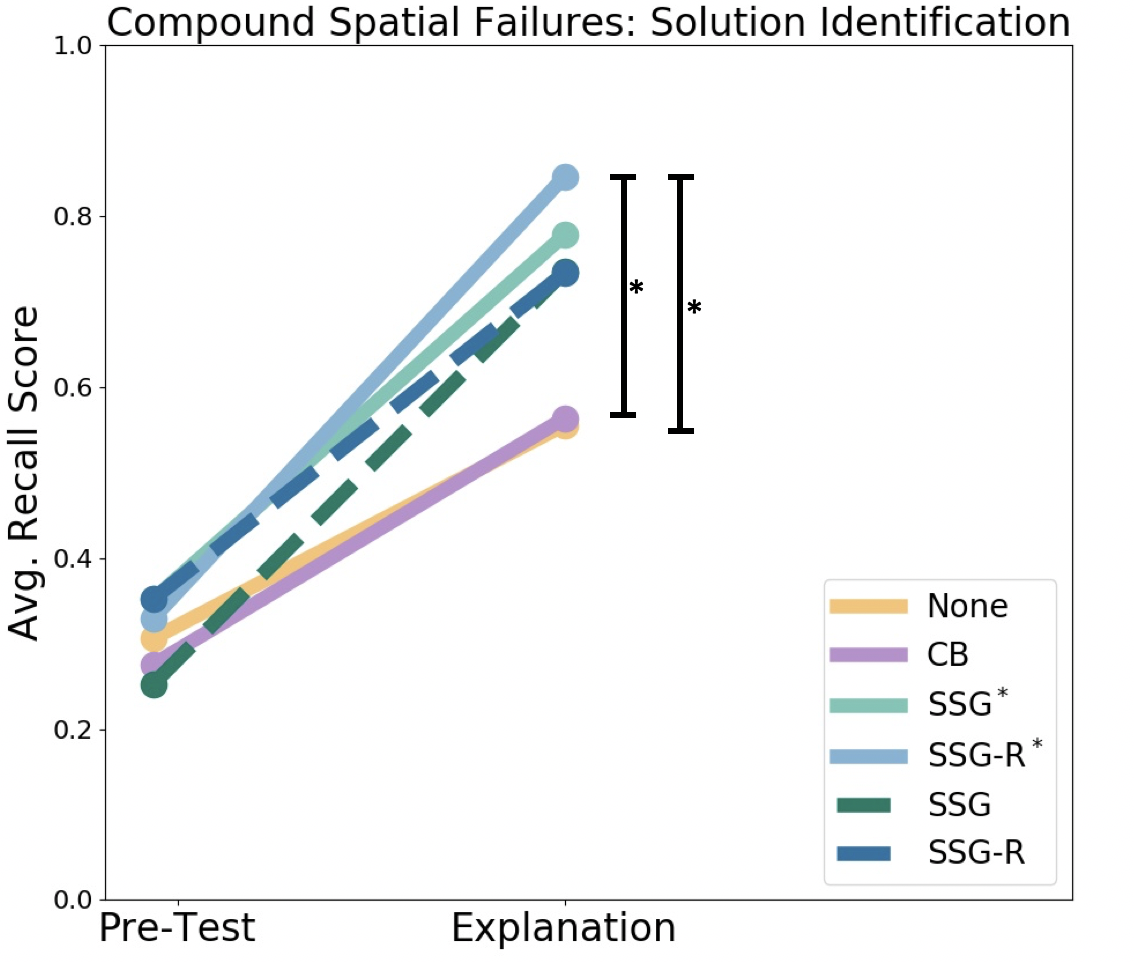}
    \caption[]%
     {{}}
     \end{subfigure}
\begin{subfigure}[b]{0.233\textwidth}
  \centering
  \includegraphics[width=\textwidth]{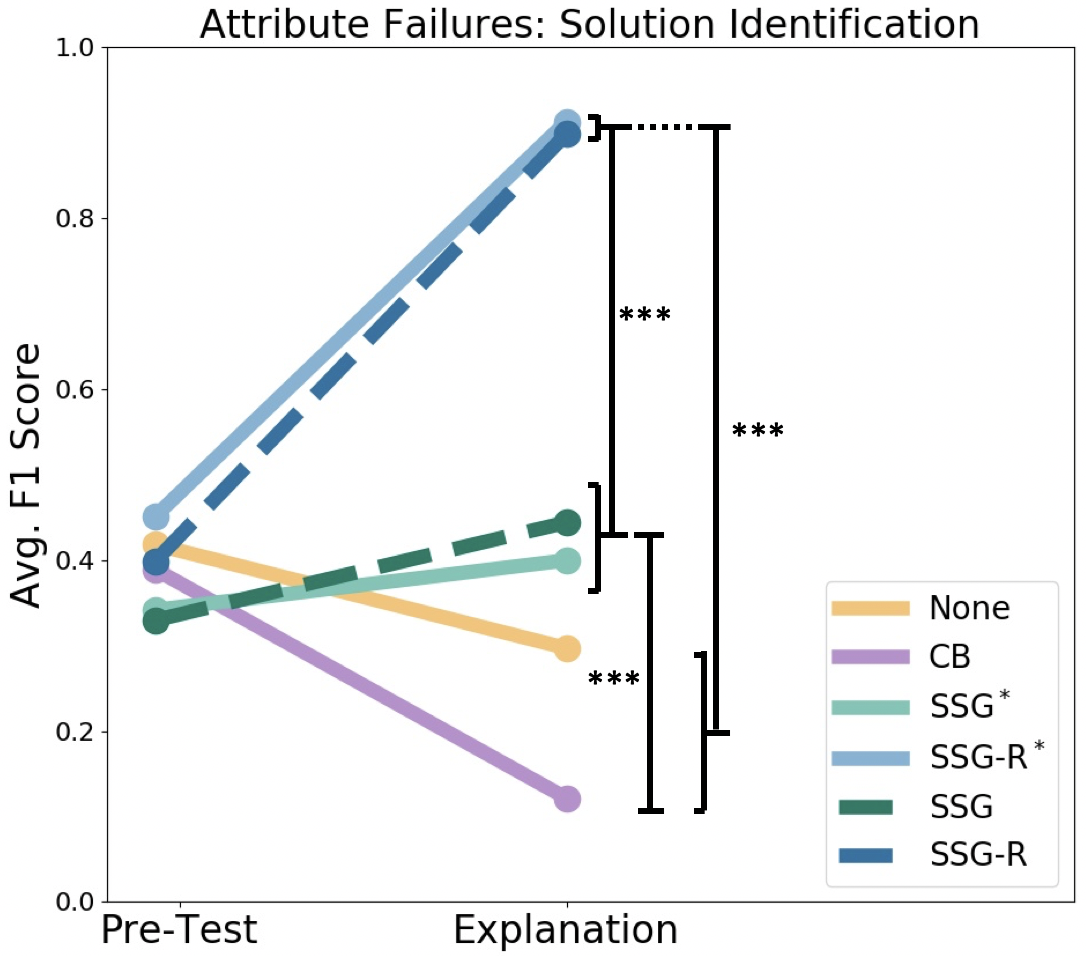}
    \caption[]%
     {{}}
     \end{subfigure}
\caption{Average F1 and Recall score for participants' solution identification across all study conditions. Statistical significance is reported as:  * p $<$ 0.05, ** p $<$ 0.01, *** p$<$ 0.001.}
\label{fig:q2-qual}
\end{figure*}

\section{Quantitative Analysis of Semantic Explanations}
\label{sec:quant-eval}
From Section \ref{sec:section-3}, we observed that semantically descriptive explanations were perceived to be more helpful for failure understanding than the existing CB explanations. In this section, we evaluate the efficacy of our model-generated explanations, $SSG$ and $SSG$-$R$ explanations in improving users' ability to identify a failure and provide assistance for recovery.  For our analyses we conducted a six-way between subjects study, with the following study conditions that differed by the type of explanation participants received:
\begin{itemize}
    \item \underline{\textit{None (Baseline)}}: Participant receives no explanation describing the cause of error. As noted by \cite{das2021explainable}, this is the standard in currently deployed robotic systems.
    \item \underline{\textit{CB (Baseline)}}: Participant receives a context-based explanation from prior work \cite{das2021explainable}.
    \item \underline{\textit{$SSG^*$}}: Participant receives a ground truth, unranked, semantically descriptive explanation. 
    \item \underline{\textit{$SSG$-$R^*$}}: Participant receives a ground truth, ranked, semantically descriptive explanation.
    \item \underline{\textit{$SSG$}}: Participant receives a model-generated, unranked semantically descriptive explanation. 
    \item \underline{\textit{$SSG$-$R$}}: Participant receives a model-generated, ranked, semantically descriptive explanation.
\end{itemize}


\subsection{Study Design}
Similar to Das et al. \cite{das2021explainable}, our user study consisted of two stages, \textit{Pre-Test} and \textit{Explanation}. 
In both stages, users were presented with images of the environment in which the robot encountered a failure  when tasked to pick up $d_{obj}$.

In the \textit{Pre-Test} stage, participants were shown 16 randomly ordered failure scenarios representing all $F_{t}$ from Figure \ref{fig:failure_types}. None of the participants were provided explanations in this stage in order to establish their initial level of error understanding. For each scenario, participants were tasked to identify the possible cause(s) of failure and suggest possible solution(s) as a remedy. 

In the \textit{Explanation} stage, participants were shown another 16 different, failure scenarios. However, participants were now provided an explanation depending on their study condition. Similar to the \textit{Pre-Test} stage, participants were tasked to identify the cause(s) of robot failure and suggest possible solution(s) as a remedy. 

\subsection{Metrics}
We evaluate participant performance using either an F1 score, or a Recall score. Since participants were allowed to select multiple answers for each question, in the case of compound spatial failure types from Figure \ref{fig:failure_types}, the quantity of false negatives are a more important measure of a participant's performance. Therefore, we analyze participants' Recall score for compound failures and F1 score for all other failure types. Specifically, we measure the difference between each participant's \textit{Pre-Test} and \textit{Explanation} F1 score or Recall score using metrics similar to Das et al. \cite{das2021explainable}:
\begin{itemize}
    \item  \underline{Failure Identification (FId)}: The ability to accurately select the correct cause(s) of failure in a scene.
    \item \underline{Solution Identification (SId)}: The ability to accurately select the actions needed to remedy the failure in a scene.
\end{itemize}  

\subsection{Participants}
We recruited 93 participants from Amazon Mechanical Turk. Participants were required to be non-experts in the domain of robotics, thus we removed three participants for scoring a 100\% accuracy on the \textit{Pre-Test} stage. The remaining 90 participants, 15 for each study condition, included 53 males and 37 females, all whom were over the age of 18 (M=39.0, SD=10.8). The task took on average 20-30 minutes and participants were compensated \$2.50.
\subsection{Quantitative Results}
The participants' failure identification (\textit{FId}), and solution identification (\textit{SId}) scores follow a normal distribution, thus we utilize a one-way ANOVA with a Tukey HSD post-hoc test to evaluate statistical significance between study conditions. 

In Figure \ref{fig:q1-qual}, we examine the average F1 score and Recall score for participants' failure identification (\textit{FId}) across the aggregated failure types as well as across each individual failure type. Overall, we see that $SSG$-$R^*$ and $SSG^*$ explanations have the highest improvement in \textit{FId} scores in comparison to the other study conditions. This indicates the effectiveness of semantically descriptive explanations in improving participants' understanding of robot failures.

When looking at the \textit{FId} scores for aggregate failures (Figure \ref{fig:q1-qual}(a)), we see  $SSG$-$R^*$, and $SSG$-$R$ explanations lead to a significant improvement in failure understanding compared to None, CB, and both $SSG$ and $SSG^*$ explanations ($p<0.001$ for all). Similar trends are  observed with single spatial failures (Figure \ref{fig:q1-qual}(b)), and attribute failures (Figure \ref{fig:q1-qual}(d)), reiterating the effectiveness of grounding explanations in the semantic information present in a scene.

For compound spatial failures (Figure \ref{fig:q1-qual}(c)), we observe that only $SSG$-$R^*$ leads to significant improvement in \textit{FId} scores. This highlights the importance of ranked semantic-based explanations in significantly improving participants' failure understanding, as well as highlights an area of improvement for our SSG model in detecting multiple failures in a scene. Furthermore, for compound failures, we also observe a strong learning effect across all study conditions. This indicates the variability in difficulty across compound spatial failures, where the \textit{Explanation} compound failure scenarios may be more visually apparent, compared to the \textit{Pre-Test} compound failure scenarios. However, this particular situation demonstrates the key reasoning for measuring the differences in performance between both stages as opposed to solely analyzing the \textit{Explanation} stage performances.

Furthermore, Figure \ref{fig:q1-qual}(d)
presents the effectiveness of $SSG$ and $SSG^*$ explanations. We observe that $SSG$ and $SSG^*$ lead to significantly improved failure understanding in comparison to None ($p<0.05$) and CB ($p<0.001$). However, $SSG$-$R$ and $SSG-R^*$ show a significantly higher rate of improvement than $SSG$ and $SSG^*$ ($p < 0.001$). This further demonstrates the benefit of ranked $SSG$-$R^*$ and $SSG$-$R$ explanations in helping participants understand the true cause(s) of robot failure. Interestingly, in Figure \ref{fig:q1-qual}(d), we observe the adverse effects CB explanations have in participants' understanding of attribute failures. Given that CB explanations can only predict a limited set of single spatial failures, when used to explain attribute failures, they cannot express the true cause of failure.

Given the trends in participants' failure identification scores, in Figure \ref{fig:q2-qual}, we also examine the average F1 score and Recall score for participants' solution identification (\textit{SId}) across all failure types. We see that participants' \textit{SId} scores closely follow the trends observed for failure identification. However, there are instances in which solution identification is harder than failure identification. For example, when analyzing single spatial failures in Figure \ref{fig:q2-qual}(b), we see that only $SSG$-$R^*$ significantly improves \textit{SId} in comparison to None, CB, $SSG^*$ and $SSG$ ($p<0.01$). Overall, we find that both $SSG$-$R^*$ and $SSG$-$R$ explanations lead to the highest \textit{SId} scores. This displays not only the importance of generating semantic-based explanations, but also the effectiveness of \textit{ranked}, semantic-based explanation that only include semantic information relevant to a failure.

\section{Conclusion And Future Work} 
\label{sec:conclusion}
In this work we have introduced a generalizable framework that autonomously captures the semantic information in a scene to explain robot pick errors to everyday users. We leverage both semantic scene graphs and pairwise ranking to develop semantically descriptive explanations that highlight the true cause of a robot failure. Our results demonstrate that ranked, semantically descriptive explanations significantly improve everyday users' ability to understand robot failures and provide assistance for fault recovery. 
Although the results are promising, there are limitations that should be addressed in future work. For example, we demonstrate that semantically descriptive explanations are useful to everyday users for understanding pick errors. It would be interesting to examine the effects of these semantic explanations in the context of other robot failures, such as navigation related errors. Additionally, while the current scene graph model can generalize over many spatial relations and object attributes, future work, in the form of additional data collection and model improvements, is required to further expand the scope of relations needed to explain a wider range of robot failures. 





\section*{ACKNOWLEDGMENT}
This material is based upon work supported by the NSF Graduate Research Fellowship under Grant No. DGE-1650044.


\bibliographystyle{IEEEtranS}
\bibliography{references}

\end{document}